\documentclass[sigconf,authorversion,nonacm]{acmart}

\usepackage{booktabs} 

\usepackage{graphicx}
\usepackage{amsmath}
\usepackage{booktabs}
\usepackage{xcolor}
\usepackage{algorithm2e}
\usepackage[noend]{algpseudocode}
\usepackage{tikz}
\usepackage{subcaption} 
\usepackage{acmart-taps}

\usepackage[capitalize]{cleveref}
\crefname{section}{Sec.}{Secs.}
\Crefname{section}{Section}{Sections}
\Crefname{table}{Table}{Tables}
\crefname{table}{Tab.}{Tabs.}

\begin{document}

\title{{L}iveNVS: Neural View Synthesis on Live RGB-D Streams}

\author{Laura Fink}
\email{laura.fink@fau.de}
\orcid{0009-0007-8950-1790}
\affiliation{%
  \institution{Friedrich-Alexander-Universit\"at Erlangen-N\"urnberg\\Fraunhofer IIS}
  \streetaddress{Cauerstr. 11}
  \country{Germany}
}

\author{Darius Rückert}
\email{darius.rueckert@fau.de}
\orcid{0000-0001-8593-3974}
\affiliation{%
  \institution{Friedrich-Alexander-Universit\"at Erlangen-N\"urnberg\\Voxray GmbH}
  \streetaddress{Cauerstr. 11}
  \country{Germany}
}

\author{Linus Franke}
\email{linus.franke@fau.de}
\orcid{0000-0001-8180-0963}
\affiliation{%
  \institution{Friedrich-Alexander-Universit\"at Erlangen-N\"urnberg}
  \streetaddress{Cauerstr. 11}
  \country{Germany}
  \postcode{91058}
}

\author{Joachim Keinert}
\email{joachim.keinert@iis.fraunhofer.de}
\orcid{0000-0003-1857-3862}
\affiliation{%
  \institution{Fraunhofer IIS}
  \streetaddress{Am Wolfsmantel 33}
  \city{Erlangen}
  \country{Germany}
  \postcode{91058}
}

\author{Marc Stamminger}
\email{marc.stamminger@fau.de}
\orcid{0000-0001-8699-3442}

\affiliation{%
  \institution{Friedrich-Alexander-Universit\"at Erlangen-N\"urnberg}
  \streetaddress{Cauerstr. 11}
  \country{Germany}
  \postcode{91058}
}



\begin{abstract}

Existing real-time RGB-D reconstruction approaches, like Kinect Fusion, lack real-time photo-realistic visualization.
This is due to noisy, oversmoothed or incomplete geometry and blurry textures which are fused from imperfect depth maps and camera poses.
Recent neural rendering methods can overcome many of such artifacts but are mostly optimized for offline usage, hindering the integration into a live reconstruction pipeline.

In this paper, we present LiveNVS, a system that allows for neural novel view synthesis on a live RGB-D input stream with very low latency and real-time rendering.
Based on the RGB-D input stream, novel views are rendered by projecting neural features into the target view via a densely fused depth map and aggregating the features in image-space to a target feature map.
A generalizable neural network then translates the target feature map into a high-quality RGB image.
LiveNVS achieves state-of-the-art neural rendering quality of unknown scenes during capturing, allowing users to virtually explore the scene and assess reconstruction quality in real-time.

\end{abstract}

\begin{teaserfigure}
  \includegraphics[width=\textwidth]{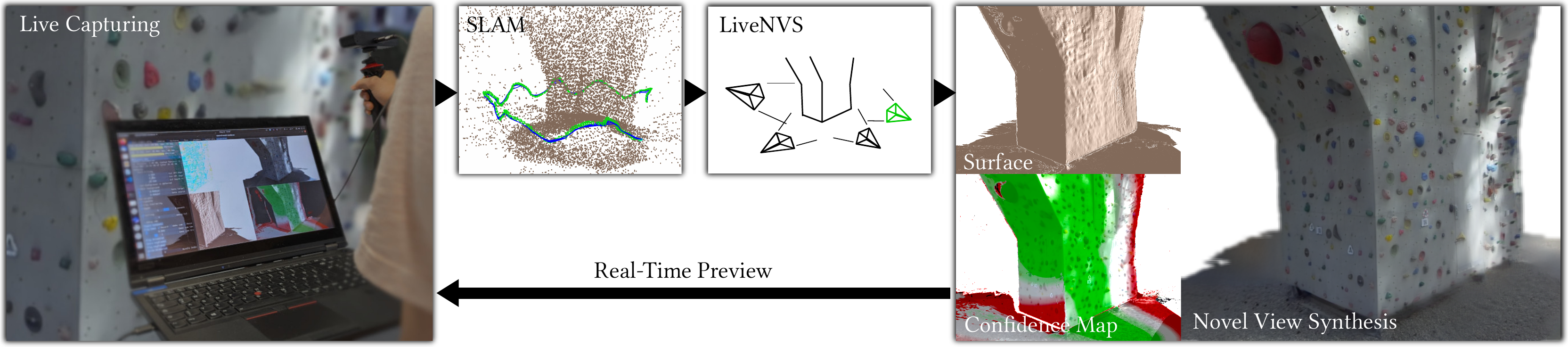}
  \caption{
  LiveNVS allows real-time novel view synthesis on a live RGB-D stream.
  The pipeline allows for live scene exploration during capturing. 
  Additionally, we can display the reconstructed surface and a confidence map of the estimated render quality, guiding the operator to identify weakly reconstructed parts of the scene.
  }
  \label{fig:teaser}
\end{teaserfigure}



\maketitle

\begin{small}
\begin{minipage}[t][][t]{\linewidth}
\vspace{0.29cm}
This is the author's version of the work. 
It is posted here for your personal use. 
Not for redistribution. 
The definitive version of record was published at SIGGRAPH Asia 2023, \url{http://dx.doi.org/10.1145/3610548.3618213}.

You can find the video that was part of the supplemental material here \url{https://youtu.be/aMbE5WAgD2k}.
The code is published on \url{https://github.com/Fraunhofer-IIS/livenvs}.

\vspace{0.4cm}
\end{minipage}
\end{small}

\section{Introduction}
\label{sec:introduction}

Live 3D reconstruction approaches, such as Kinect Fusion \cite{newcombeKinectFusionRealtimeDense2011}, take as input a stream of 3D sensor data plus an RGB stream, and provide an instant reconstruction of the seen geometry.
They also show colored previews of the reconstruction during the capturing process.
However, for practical application in VR, AR, telepresence and others, the rendering quality of the previews is not sufficient, mostly due to lacking texture detail and resolutions, as well as missing view-dependent effects.

\begin{table*}[]
\small
    \centering
        \caption{ Applicability of comparable NVS-methods for real-time reconstruction and live preview regarding their ability to (i) provide interactive reconstruction and (ii) rendering. The method can handle (iii) fast growing datasets and can (iv) instantaneously incorporate updates on camera poses from an concurrent calibration process (SLAM). (v) The pipeline implements mechanisms to be less vulnerable to an imperfect capturing process (hand-held, no studio setup) and camera poses (reliance on very accurate MVS calibration). (vi) Plausible results even for sparsely captured scene regions (e.g. no cloudy artifacts). Mip-NeRF 360 \cite{barron2022mipnerf360}, Instant Neural Graphics Primitives \cite{muller2022instant}, Volumetric Bundle Adjustment~(VBA)~\cite{clarkVolumetricBundleAdjustment2022}, NeRF-SLAM \cite{rosinolNeRFSLAMRealTimeDense2022}, MVS-NeRF \cite{chenMVSNeRFFastGeneralizable2021}, Stable View Synthesis~(SVS)~\cite{rieglerStableViewSynthesis2021}, Real-Time Novel View Synthesis With Forward Warping~(FWD) \cite{caoFWDRealTimeNovel2022}, LiveNVS (ours).}
\begin{tabular}{ll|c|c|c|c|c|c|c|c}
      &                                               & MIP-NeRF & Instant & VBA & NeRF- & MVS- & SVS & FWD & LiveNVS \\
      & Capability                                    & 360      & NGP     &     & SLAM  & NeRF &     &     & (ours)  \\ \hline
(i)   & Live reconstruction                           & -        & +       & ++  & ++    & -    & -   & ++  & ++      \\
(ii)  & Real-time rendering                           & -        & +       & +   & +     & -    & +   & +   & ++      \\
(iii) & Growing datasets                              & -        & +       & +   & ++    & ++   & -   & ++  & ++      \\
(iv)  & Fast response to pose corrections             & -        & +       & +   & +     & ++   & -   & ++  & ++      \\
(v)   & Robust to imperfect capturing or camera poses & -        & -       & -   & +     & -    & +   & -   & +       \\
(vi)  & Sparse input views                            & -        & -       & -   & -     & ++   & -   & +   & +      
\end{tabular}
    \label{tab:properties}
\end{table*}
 
\begin{figure}
    \centering
    \includegraphics[width=\linewidth, height=3.1cm]{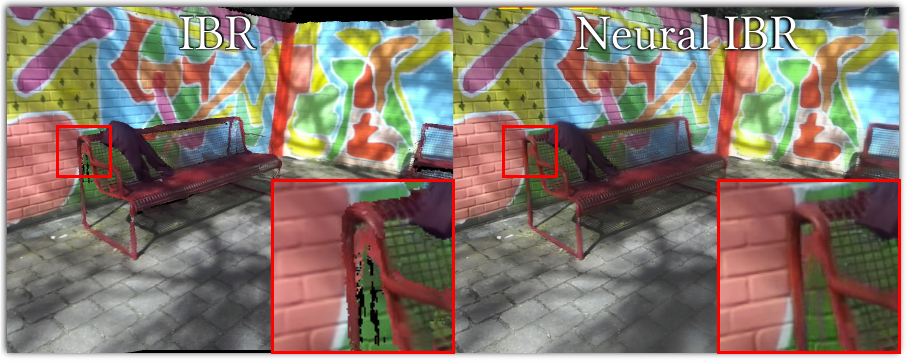}
    \caption{ 
    Conventional IBR (left) generates sharp results but introduces artifacts at incorrect or missing geometry. Neural IBR methods (like ours) reduce these artifacts without losing image sharpness (right).
    }
\label{fig:ff_neural}
\end{figure}

On the other hand, recent improvements in deep neural network architectures have led to a resurgence of interest in image-based rendering (IBR) and novel view synthesis (NVS), as common artifacts can be rectified effectively by these methods (see Figure \ref{fig:ff_neural}).
State-of-the-art IBR and NVS approaches \cite{rieglerStableViewSynthesis2021,barron2022mipnerf360} are able to render novel views that are hard to differentiate from real photographs.
However, these methods are not directly applicable to live 3D reconstructions for multiple reasons, in particular:
\begin{itemize}
\item high preprocessing times / scene-specific optimizations
\item non-interactive render times
\item reliance on dense and/or high-quality input data
\item no support of growing datasets and late pose corrections (e.g. loop closure)
\end{itemize}
This is not only detrimental for on-site 3D reconstruction, because a user does not get instantaneous feedback whether they have captured sufficient images, but also prevents latency critical applications such as interactive telepresence~\cite{jahromiYouDriveMe2020, tanHumanRobotCooperationBased2020}.
Tab.~\ref{tab:properties} gives an overview of recent related neural novel view synthesis approaches and their properties relevant for real-time reconstruction from a live RGB(-D) stream.
 
In this work, we close this gap and present an integrated 3D reconstruction and neural rendering pipeline LiveNVS (see Fig.~\ref{fig:teaser}), that overcomes the above mentioned limitations.
LiveNVS is executed on live RGB-D streams, and delivers instant renderings of novel views in a quality that is on par with comparable state-of-the-art NVS approaches.
Our major contribution is an image-space fusion algorithm that aggregates a varying number of warped neural features in a novel view.
After that, a neural network translates the feature map of the novel view to a photorealistic RGB image.
Since only a few RGB-D images instead of a global scene representation are required, our approach can naturally support growing datasets and is very time and memory efficient.
Furthermore, we support continuous scene and pose updates during the capturing process, which is relevant for e.g. loop closure.

In short, our contributions are:

\begin{itemize}
\item A real-time novel view synthesis pipeline with neural rendering capabilities that runs on a live RGB-D stream as input
\item A novel differentiable image-space forward-warping and fusion method for neural image features which allows for camera pose updates like loop closures during capturing
\item Multiple other features, such as robust view selections and motion blur defeating strategies, which greatly improve our quality and can easily be integrated in common pipelines
\item An open source implementation available on \url{https://github.com/Fraunhofer-IIS/livenvs}.

\end{itemize}

\section{Related Work}
\label{sec:related_work}

Synthesis of novel views from a sequence of input RGB-D input requires to
(i) compute the extrinsic and intrinsic camera parameters, 
(ii) fuse several or all input frames
into a common representation, and
(iii) synthesize an image representing the scene or object from the desired target camera pose.

\paragraph{Camera parameters}
Immediate visual feedback on the current digitization status of the scene (or the object) requires all the previous steps to run in real-time with low latency.
This excludes many of the methods typically used for pose estimation.
For instance, the commonly used COLMAP~\cite{schonbergerPixelwiseViewSelection2016}
has no real-time capabilities and requires all input views before any geometry reconstruction can be started, thus preventing early user feedback.

In contrast, simultaneous localization and mapping~(SLAM) allows for instant feedback about the camera trajectory and recorded surfaces \cite{mur-artalORBSLAM2OpenSourceSLAM2017, macariobarrosComprehensiveSurveyVisual2022}.
Since frames are added sequentially, camera extrinsics can contain significant errors and even drift, and can be updated in retrospective by global bundle adjustment or loop closure.

\paragraph{Fusion into a common representation}

After successful determination of the camera parameters, the input frames may be combined into a common representation.
Multi-view stereo algorithms \cite{griwodzAliceVisionMeshroomOpensource2021,zhuDeepLearningMultiView2021a,furukawaMultiviewStereoTutorial2015,schonbergerPixelwiseViewSelection2016,wangPatchmatchNetLearnedMultiView2020} combine the input views into point-clouds or textured meshes that can then be used to render a novel view.
However, those methods are time consuming.
(Truncated) signed distance fields \cite{newcombeKinectFusionRealtimeDense2011, daiBundleFusionRealtimeGlobally2017,koestlerTANDEMTrackingDense2022} and methods based on surface elements (surfels) \cite{whelanElasticFusionRealtimeDense2016, ruckertFragmentFusionLightWeightSLAM2019, schopsSurfelMeshingOnlineSurfelBased2020} can be implemented in real-time, but they lack photorealistic rendering.

\paragraph{View synthesis based on scene dependent optimization}

Use of neural rendering methods \cite{tewariStateArtNeural2020} can boost rendering quality.
\citet{thiesDeferredNeuralRendering2019} optimize an object-specific neural texture for a surface mesh thus being able to reproduce view-dependent effects and fixing erroneous silhouettes in image space.
\citet{aliev_neuralpointbasedgraphics_2020}, \citet{kopanasPointBasedNeuralRendering2021} as well as \citet{ruckert_adop_2021} use point clouds as proxy and optimize a descriptor per point.

Alternatively, discrete voxel grids can be used \cite{muller2022instant, sunDirectVoxelGrid2021,clarkVolumetricBundleAdjustment2022}. 
Rendering can be fast \cite{liRTNeRFRealTimeOnDevice2022,linEfficientNeuralRadiance2022,espositoKiloNeuSVersatileNeural2022,reiserMERFMemoryEfficientRadiance2023} to even work on mobile devices \cite{caoRealTimeNeuralLight2022}.
While real-time reconstruction is significantly more difficult, careful optimization and camera parameter refinement permits fast capture and view synthesis \cite{haitzCombiningHoloLensInstantNeRFs2023,mullerInstantNeuralRadiance2022,jiangInstantNVRInstantNeural2023,clarkVolumetricBundleAdjustment2022,rosinolNeRFSLAMRealTimeDense2022}.
Other approaches demonstrate their application on video data with dynamic content \cite{songNeRFPlayerStreamableDynamic2022,liStreamingRadianceFields2022,liDynIBaRNeuralDynamic2023}.
Nevertheless, many explicit approaches are computation intensive and memory-hungry if they store voxel grids of complexity $O(n^3)$.
While some lower memory demands using hash grids~$(O(n^2)$~\cite{muller2022instant}, implicit representations like Neural Radiance Fields (NeRFs)~\cite{mildenhallNeRFRepresentingScenes2020a,barron2022mipnerf360} can reduce the required memory even further, albeit at the expense of huge training times.
\citet{xieNeuralFieldsVisual2022} provide an extensive overview on this topic.

As all above methods come at the cost of scene specific optimization or heavy pre-preprocessing (camera calibration, proxy geometry computation), many only support limited resolution \cite{rosinolNeRFSLAMRealTimeDense2022} or need massive hardware requirements \cite{clarkVolumetricBundleAdjustment2022}.

\paragraph{Generalizing neural networks for view synthesis}

Scene specific optimization can be avoided by using generalizing neural networks trained only once.
Consequently, novel view synthesis promises to take less time from capturing to the first available frame.

Hedman et al.~\shortcite{hedmanDeepBlendingFreeviewpoint2019} acquire pixel-wise blending weights for image-based rendering.
Riegler and Koltun~\shortcite{rieglerFreeViewSynthesis2020} add an additional CNN that encodes source views into neural feature maps prior to warping, then use a decoder to ouput images and blending weights. 
Similarly, there exist several other approaches that integrate encoding networks to the pipeline~\cite{rakhimovNPBGAcceleratingNeural2022, rieglerStableViewSynthesis2021, jainEnhancedStableView2023, wangIBRNetLearningMultiView2021a}, yielding an encoding-warping-aggregation-decoding or encoding-warping-decoding-aggregation procedure.
These encoders predict feature maps for source images, thus allowing to acquire neural descriptors in a generative fashion.
When trained on sufficiently big datasets, such pipelines can be scene-agnostic and generalize well also for unseen objects. 
These principles are applied to point-based rendering \cite{rakhimovNPBGAcceleratingNeural2022}, signed distance functions \cite{bergmanFastTrainingNeural2021} and proxy meshes \cite{rieglerStableViewSynthesis2021, jainEnhancedStableView2023}.
\citet{chenMVSNeRFFastGeneralizable2021} alternatively predict NeRFs.
\citet{wangIBRNetLearningMultiView2021a} predict densities along rays of the target image using a ray transformer that is given features from encoded source images. 

However, despite avoiding scene specific training, none of those methods work in real-time on a continuous stream of input frames due to heavy pre-processing, complex network architectures or slow rendering.
Following these observations, we identify a significant gap regarding interactivity within the state of the art.

\paragraph{Real-time neural scene reconstruction and view synthesis}

To cope with those limitations, several works specifically address the challenge of an end-to-end real-time pipeline.
This can be achieved using global neural fields~\cite{sucarIMAPImplicitMapping2021, lionarNeuralBloxRealTimeNeural2021, ortizISDFRealTimeNeural2022, zhuNICESLAMNeuralImplicit2021}
or separate neural representations for depth map fusion~\cite{wederNeuralFusionOnlineDepth2020, choeVolumeFusionDeepDepth2021}. 

However, all of these methods focus on geometry reconstruction and struggle in creating photo-realistic high quality renderings due to memory intensive data structures or long rendering times.
Cao et al.~\shortcite{caoFWDRealTimeNovel2022} tries to overcome these difficulties by proposing an encoding-warping-aggregation-decoding scheme for real-time novel view synthesis.
The individually warped views are merged by a transformer network.
By these means, it is the closest method to our approach.
However, due to the transformer complexity, the supported image resolutions are small.
Moreover, the evaluation is restricted to DTU \cite{jensenLargeScaleMultiview2014} and ShapeNet \cite{changShapeNetInformationRich3D2015} datasets, excluding many real-world challenges such as imperfect camera parameters.

\paragraph{Our contributions}

We combine photo-realistic neural rendering with real-time 3D reconstruction.
To avoid a lack of memory for large-scale scenes, we resign to any global data structure
and rely on the principles of (depth) image-based rendering~((D)IBR). 
We augment the fusion algorithm by \citet{ruckertFragmentFusionLightWeightSLAM2019} with a refined neural weighting scheme to be robust against noisy depth maps and inaccurate lens undistortions. 
A light-weight implementation achieves interactive results using a fast and screen-space neural descriptor fusion algorithm.
Thus, we allow for free view point synthesis already during capturing, which is a feature underrepresented in the field of NVS.

\newcommand{\enc}{$e$\xspace}
\newcommand{\dec}{$d$\xspace}
\newcommand{\feat}{$f$\xspace}
\newcommand{\featn}{$f_n$\xspace}
\newcommand{\featt}{$f_t$\xspace}
\newcommand{\col}{$c$\xspace}
\newcommand{\tgt}{$t$\xspace}
\newcommand{\srcm}{s\xspace}
\newcommand{\src}{$\srcm$\xspace}
\newcommand{\srcn}{$\srcm_n$\xspace}
\newcommand{\depthn}{$d_n$\xspace}
\newcommand{\depth}{$d$\xspace}
\newcommand{\Kn}{$K_n$\xspace}
\newcommand{\Rn}{$R_n$\xspace}
\newcommand{\transn}{$t_n$\xspace}
\newcommand{\K}{$K$\xspace}
\newcommand{\R}{$R$\xspace}
\newcommand{\trans}{$t$\xspace}
\newcommand{\svg}{{camera local mesh}\xspace}
\newcommand{\svgs}{{camera local meshes}\xspace}

\begin{figure*}[t]
\centering
\includegraphics[width=\linewidth]{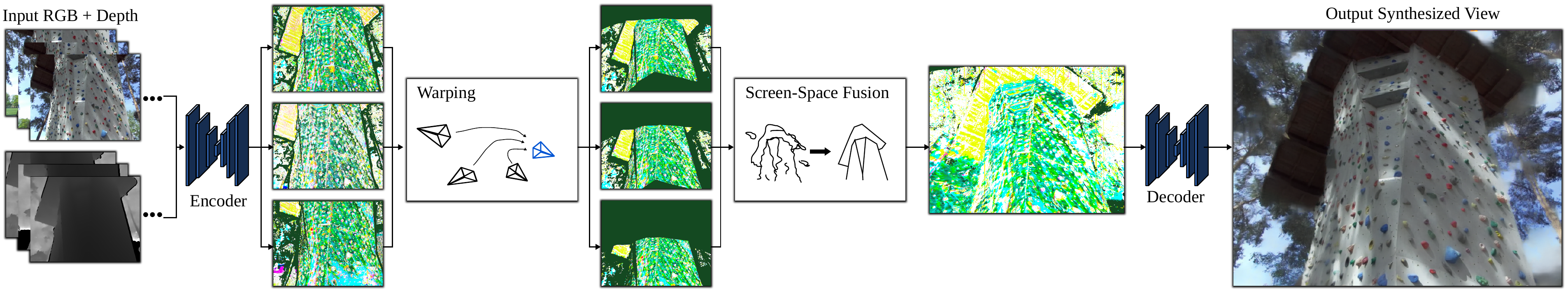} 
\caption{ 
Overview of our Pipeline. 
The input is a stream of RGB-D images, which are encoded by a neural network.
After that, the features are warped to a target frame and fused in screen-space.
Finally, a lightweight decoder network translates the features to a photo-realistic color image.
}
\label{fig:pipeline}
\end{figure*}

\section{Method}
\label{sec:method}

Fig.~\ref{fig:pipeline} shows an overview of our neural rendering pipeline.
The input is a set of RGB-D images and the corresponding camera poses, for example, from a real-time SLAM system (Sec.~\ref{ssec:slam}).
We carefully select suitable images as keyframes (Sec.~\ref{ssec:keyframes}) considering overlap between keyframes and motion blur.
To synthesize a novel view, several nearby keyframes are selected (Sec.~\ref{ssec:view_selection}), encoded to latent space by a neural network (Sec.~\ref{ssec:encoding}), and warped to the target image (Sec.~\ref{ssec:warp}).
The warped feature vectors are then fused by our novel screen-space feature fusion function (Sec.~\ref{ssec:fuse}) using tailored feature weighting in order to achieve stable results.
We also present a deferred warping mode that further increases stability  (Sec.~\ref{ssec:deferredwarping}).
Finally, a decoder network reconstructs the final RGB image of the novel view (Sec.~\ref{ssec:decoding}), where temporal stability is improved by including information from prior frames.

\subsection{SLAM}
\label{ssec:slam}

The first step in our pipeline is a simultaneous localization and mapping (SLAM) system that computes a 6-DoF pose of each input image.
We are using Snake-SLAM \shortcite{9476760} due to its high efficiency and accurate pose estimates.
In the SLAM-system previous key frames are continuously optimized by local bundle adjustment and global loop closure operations.
Our surface fusion (see Sec.~\ref{ssec:fuse}) can handle such late updates, because we have no global model and the 3D information is stored in the local space of the key frames, and thus does not change when the key frame's pose is modified.

\subsection{Key Frame Selection from Input Stream}
\label{ssec:keyframes}

To make the input video stream more manageable for rendering, we derive a set of keyframes. 
Our keyframe selection mechanism is similar to that of other SLAM-systems such as ORB-SLAM \cite{mur-artalORBSLAM2OpenSourceSLAM2017} but we select frames based on a motion score.
The score is based on the observation that effects like motion blur or rolling shutter distortions degrade input data.
It is calculated from the per-pixel motion vector using consecutive poses and depthmaps.
The average of the length of the motion vectors across the image constitutes the score per frame.
We evaluate this score over a 1\,s window, selecting the optimal frame. 
Despite neglecting shutter speed, our approach's efficacy is validated in Sec.~\ref{sec:results}.

\subsection{View Selection for Rendering}
\label{ssec:view_selection}
To achieve real-time frame rates and counteract over-smoothing the final result, we select the $N$ (in our examples $N=15$ if not noted otherwise) best nearby key frames for the target image.
A careful selection is essential to avoid large holes and blurry edges. 
To that end, we follow the idea of \citet{hedmanDeepBlendingFreeviewpoint2019} by computing the coverage of a source frame in the target view but extend it in two ways.
First, our coverage is calculated on a per-tile basis, which guarantees that each tile of the target is seen at least once.
Second, the candidate views of each tile are sorted by the projected weighting scheme (see Sec.~\ref{ssec:weights}) to prefer candidates with good warping properties.
Hence, the algorithmic complexity depends on the number of available source views.

\subsection{Encoding}
\label{ssec:encoding}
After view selection, the $N$ best RGB-D source images are transformed to latent space by a deep neural network.
The used architecture is based on an adapted ResUNet with the encoder part being pre-trained on the Image-Net dataset~\cite{rieglerStableViewSynthesis2021}.
It receives the linear depth as extra auxiliary input.
The main output is a 4D feature vector for each pixel, and additionally a confidence value that self masks the prediction, e.g. if depth and color information do not match.
We concatenate the RGB-D input with the 4D output to generate the 8D feature vector used for warping. 
Since the encoded views are usually required multiple times, they get cached in a least-recently-used buffer.
To limit the variance in rendering time, we set a fixed number of encoding passes per frame resulting in a temporal build-up of the target view.

\subsection{Warping}
\label{ssec:warp}

To warp the latent space features maps from the source to the target view, we follow a forward warping approach that requires no global scene model.
The forward warping is implemented by triangulating the depth maps (with occlusion edges removed) and rendering them to the target view using the standard rasterization pipeline.
The warped triangles are then immediately blended in the target view, as described in the next section.
This implementation is computationally very efficient and does not require dynamic memory allocation during rasterization.

\subsection{Screen-Space Surface Fusion}
\label{ssec:fuse}

From the warping stage, we obtain triangles textured with features from the input images, which are rasterized into the target view.
Our fusion algorithm, which is inspired by the work of~\citet{ruckertFragmentFusionLightWeightSLAM2019}, incrementally combines these fragments to obtain a final single neural color vector.

For each output pixel, we keep track of the current linear depth $d$, weight $w$, and the neural feature $f$.
We take into account that depth accuracy of 3D sensors decreases with distance, which we describe using a depth error function $\Delta(d)$, which is modelled as inversely proportional to a quadratic polynomial \cite{mallickCharacterizationsNoiseKinect2014}:
\begin{equation}
    \Delta(d) = \dfrac{1}{ad^2 + bd + c},
    \label{eq:w_d}
\end{equation}
where weights $a,b,c$ have to be selected according to Mallick et al.~\shortcite{mallickCharacterizationsNoiseKinect2014}.

If a new fragment with $d_f$, $w_f$, and $f_f$ is rendered to the screen the per pixel values are updated based on one of the following cases:
\begin{enumerate}
    \item If $d_f$ < $d - \Delta(d)$
    \begin{itemize}
        \item[$\rightarrow$]  New fragment belongs to a new surface in front of the current surface.
        \item[$\rightarrow$] Overwrite current pixel values
    \end{itemize}
    \item If $d_f$ > $d + \Delta(d)$
    \begin{itemize}
        \item[$\rightarrow$]  New fragment is behind the current surface
        \item[$\rightarrow$] Discard fragment
    \end{itemize}
    \item Else
    \begin{itemize}
        \item[$\rightarrow$] New fragment is near the current surface
        \item[$\rightarrow$] Fuse fragment values into pixel
    \end{itemize}
\end{enumerate}

In the third case, the fragment is fused into the pixel updating the current pixel values using an incremental averaging scheme:
\begin{align*}
    \alpha &\leftarrow w / (w + w_f) \\
    w &\leftarrow   w + w_f\\
    d &\leftarrow \alpha d + (1-\alpha)d_{f}\\
    f &\leftarrow \alpha f + (1-\alpha)f_{f}
\end{align*}

\subsection{Feature Weighting}
\label{ssec:weights}

The final feature descriptor of each pixel is a weighted sum of the contributing source features.
The weight for each source feature should closely match the image quality at that 3D point, less reliable and biased input should be weighted lower.
For example, if the source view is close to the surface and its incident angle is similar to the target angle, a large weight should be used.
If the source view is far away or observes the surface from a different angle, a small weight should reduce the influence because we expect an erroneous and blurry warping procedure.
To that end, we apply a weighting consisting of three parts:
\begin{equation}
w_f = (w_d w_v w_i)^5,
\end{equation}

where $w_d$ is the depth accuracy, $w_v$ the view direction weight, and $w_i$ a vignetting coefficient.
To compute depth accuracy, we use the previously defined depth error function:
\begin{equation}
    w_d = \Delta(d_f).
    \label{eq:w_d}
\end{equation}

The viewing direction weight $w_v$ depends on the angle between the fragment's target~~$v_t$ and source view direction~$v_s$.
Using the dot product, the weight can then be defined as:
\begin{equation}
w_v = \max(0,v_s \cdot v_t).
\end{equation}

Finally, the vignetting coefficient $w_i$ downweights pixels with their distance $c$ to the image center:
\begin{equation}
w_i = 1 - \frac{c}{c_{\mathrm{max}}}.
\end{equation}
This weight has multiple purposes: accounting for imperfect lens undistortion and degrading depth accuracy with distance to the center.
Additionally, sharp transitions at boundaries of the blended input images are smoothed out.

\begin{figure}
    \centering
    \includegraphics[width=\linewidth, height=7.2cm]{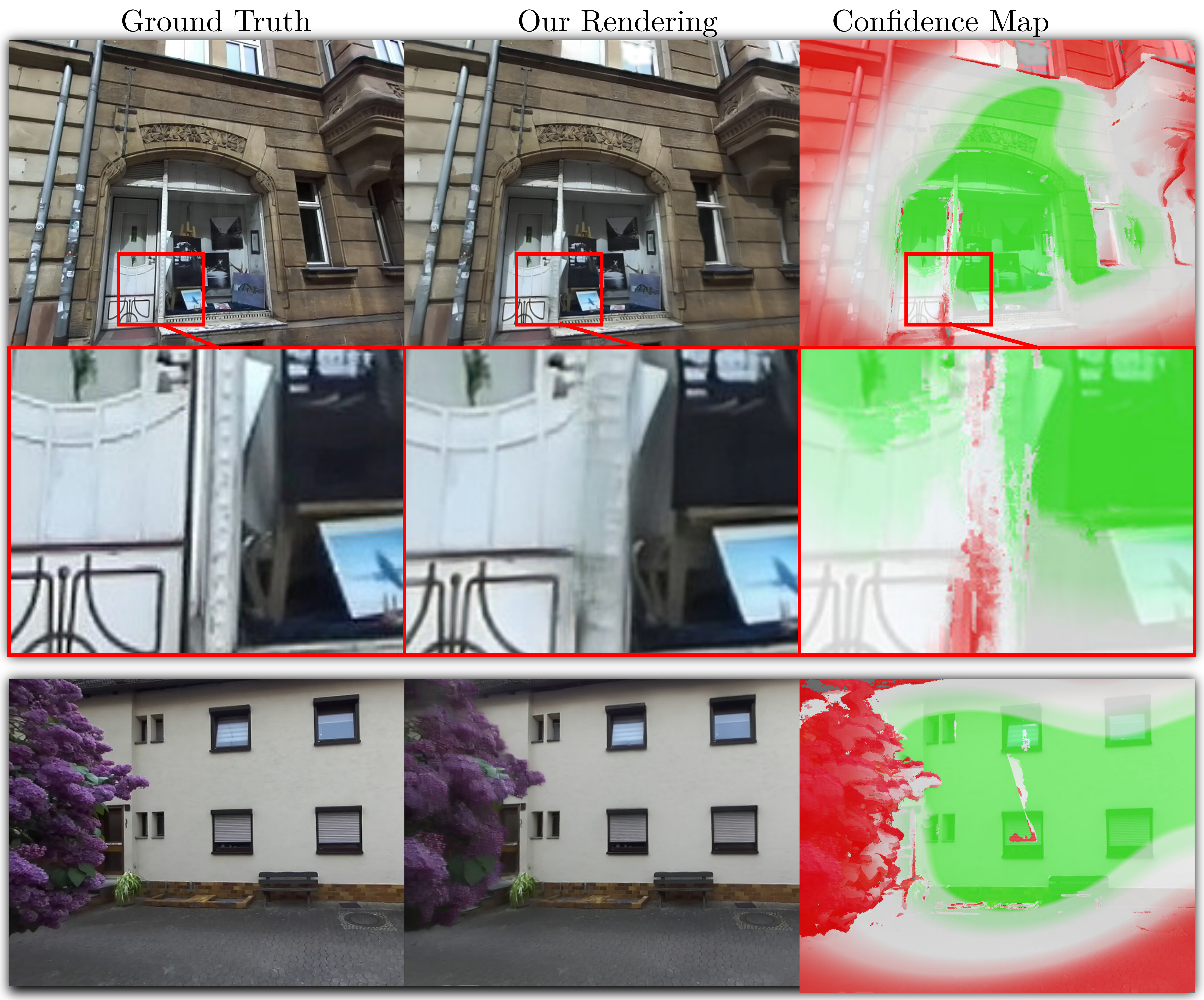}
    \caption{ 
    Ground truth, our neural rendering, and our predicted reconstruction confidence map. Quality ranging from green (good) over white (normal) to red (bad).
    }
\label{fig:heatmap}
\end{figure}

\paragraph{Reconstruction Guidance}
The average of the per-pixel weight~$w$ can in addition be beneficial as guidance for the expected image quality.
When overlayed on the rendering (see Fig.~\ref{fig:heatmap}), the operator can be guided to undersampled regions of the scan. 

\begin{figure*}[]
\centering
\includegraphics[width=0.498\linewidth]{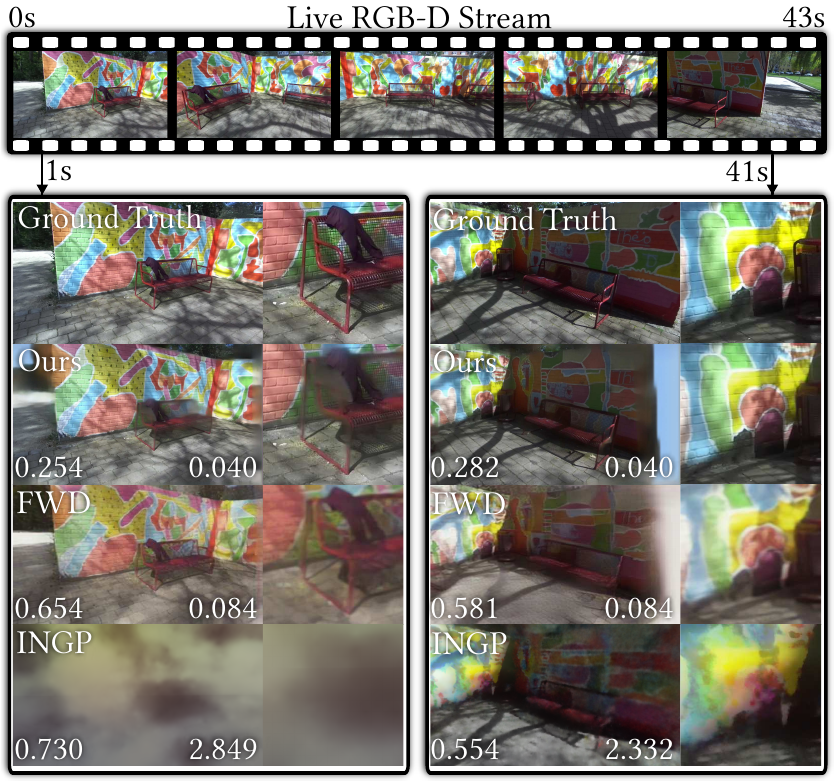} 
\hfill
\includegraphics[width=0.498\linewidth]{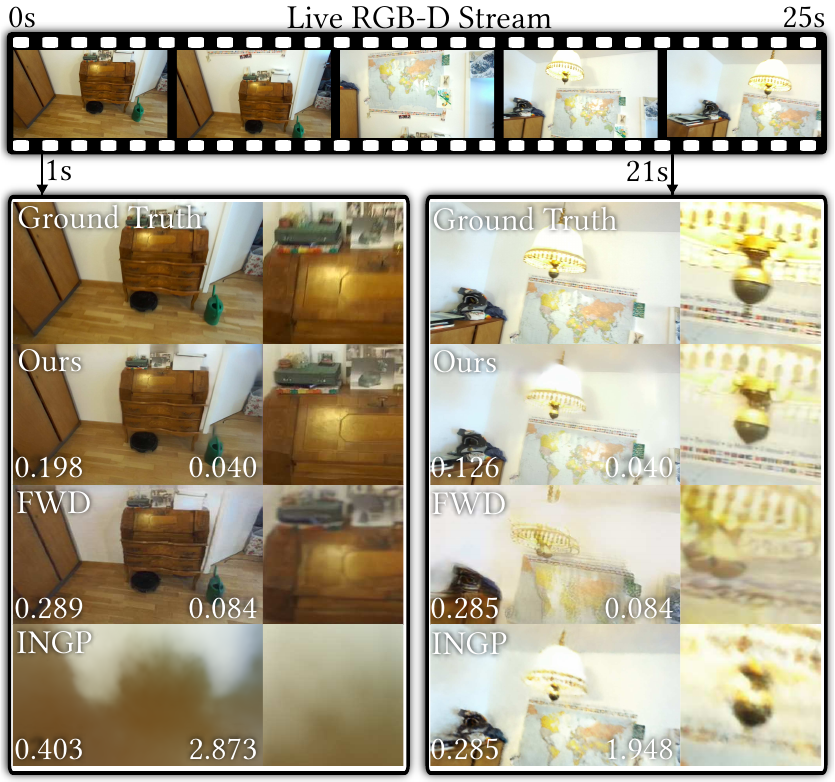} 

\caption{ 
Live novel view synthesis results shown on an outdoor and indoor scene. 
Ours and FWD output novel views using up to 6 source views. 
Values on the lower left corner indicate LPIPS\,$\downarrow$ score and values on the right the render times in ms measured on a Nvidia RTX 2080Ti at a resolution of 1280$\times$720 (*FWD only renders at resolution  400$\times$300.) INGP was optimized for the time span from capturing start up to the last source view previous to the novel view using the full trajectory.
Our system outperformed FWD and INGP in regards of LPIPS and render times in this scenario.
}
\label{fig:instant_nvs}
\end{figure*}

\subsection{Deferred Warping Mode}
\label{ssec:deferredwarping}

We have observed that the fused depth images are strongly influenced by the selection of source views, resulting in temporal instability in the output. To address this issue, we introduce a two-stage \textit{deferred warping} mode in our pipeline to separate the surface and feature fusion processes.
In the first stage, the depth map of the target view is computed as described earlier. Following that, the features are aggregated by sampling from the source views using the estimated reprojected depth value. Since the depth of the surface is already known during the reprojection, we update $w_d$ to reflect the distance between the estimated and projected depth values:
\begin{equation}
    w_d = 1- (|d-d_f|/\Delta(d))^2,
\end{equation}

The deferred warping mode of our pipeline offers a notable advantage: it allows us to leverage a larger number of contributing depth maps for surface fusion, while at the same time reducing the number of input images for feature aggregation.
This approach thus enhances the temporal stability of renderings while having minimal impact on performance, because the time required for depth map fusion is negligible compared to feature encoding and caching.

\subsection{Decoding with Temporal Feedback Loop}
\label{ssec:decoding}

In the final stage of our pipeline, we convert the fused feature map into an RGB image, as shown in Fig.~\ref{fig:pipeline}.
To do so, we employ a decoder that builds upon the well-established UNet architecture, extended by a feedback loop to improve temporal stability.
For each resolution level, the previous frame's intermediate features are linearly blended with the intermediate features of the current frame. 

Since some of our training datasets do not include video data, we mimic previous frames by applying minor spatial transformations to the current frame during training. This approach enables us to train a temporally stable decoder network even when we only have a limited number of images with significant differences.
Note that we do not reproject the previous frame's features, instead the decoder network is trained to account for this.

\section{Results}
\label{sec:results}

In this section, we show results produced by our method.
Our experimental setup and procedure is described in Sec.~\ref{ssec:data_and_train} and the supplemental material. 
Our results are compared to other methods with respect to quality and view synthesis time.
We provide extensive ablation studies on key frame selection, view extrapolation, late pose updates, source of depth maps and our view selection.


\aptLtoX[graphic=no,type=html]{
\begin{table*}
\footnotesize
    \centering
    \caption{Quantitative comparison given as LPIPS$\downarrow$ score. The best two methods are printed in bold.}
\begin{tabular}{@{}l|c|c|c@{}}
       & \textsc{Scannet}             & \begin{tabular}[c]{c}\textsc{Motor-}\\ \textsc{cycle} \end{tabular}          & \textsc{Sofa}                \\ \hline
Ours   & \textbf{0.274}      & 0.303               & \textbf{0.234}      \\
FWD    & 0.428               & 0.451               & 0.295               \\
SVS    & 0.316               & \textbf{0.247}      & 0.237               \\
IBRNet & \textbf{0.248}      & \textbf{0.179}      & \textbf{0.149}      \\
NPBG++ & 0.396               & 0.500               & 0.366               \\
DIBR   & 0.325               & 0.290               & 0.313               \\
    \end{tabular}
    \label{tab:big_eval}    
\end{table*}
\begin{table*}
\footnotesize
    \centering
        \caption{Time consumption for different methods. Preprocessing includes training time and the time to generate the proxy geometry. (*FWD only renders at resolution  400$\times$300.) }
    \begin{tabular}{l|ccc}
        Method & Proxy Geometry & Preprocessing & Rendering Time\\ 
        \hline
        Ours & depth maps & < 1\,s & $\sim$ 46\,ms\\ 
        FWD & depth maps & < 50\,s & $\sim$ 84\,ms*\\ 
        SVS & mesh & $\sim$ 4\,h & $\sim$ 3\,s\\ 
        IBRNet & {---} & < 1\,min & $\sim$ 2\,min\\ 
        NPBG++ & point cloud & $\sim$ 1\,h & $\sim$ 100\,ms \\ 
        DIBR & depth maps & < 1\,min & $\sim$  5\,ms\\ 
        ADOP & point cloud & $\sim$ 8\,h & $\sim$ 15\,ms \\ 
        Instant-NGP & volumetric hash & $\sim$ 2\,min & $\sim$ 2\,s \\ 
    \end{tabular}
    \label{tab:time}
\end{table*}
\begin{table*}
\footnotesize
    \centering
        \caption{Keyframe selection results on \textsc{Scannet} using frames that passed our selection ({\small s}) or using every 20th view ({\small mod20}) during training. }
    \begin{tabular}{l|c}
        Training Views & LPIPS  $\downarrow$  \\ 
        \hline 
        Ours\textsubscript{s} & 0.2705 \\ 
        Ours\textsubscript{mod20} & 0.3401 \\ 
        SVS\textsubscript{s} & 0.2673 \\ 
        SVS\textsubscript{mod20} & 0.3162 \\ 
    \end{tabular}
    \label{tab:scannet_motion_blur}
\end{table*}}{
\begin{table*}
\footnotesize
\begin{minipage}[t][][t]{0.26\linewidth}
    \centering
    \caption{ Quantitative comparison given as LPIPS$\downarrow$ score. The best two methods are printed in bold.}
\begin{tabular}{@{}l|c|c|c@{}}
       & \textsc{Scannet}             & \begin{tabular}[c]{c}\textsc{Motor-}\\ \textsc{cycle} \end{tabular}          & \textsc{Sofa}                \\ \hline
Ours   & \textbf{0.274}      & 0.303               & \textbf{0.234}      \\
FWD    & 0.428               & 0.451               & 0.295               \\
SVS    & 0.316               & \textbf{0.247}      & 0.237               \\
IBRNet & \textbf{0.248}      & \textbf{0.179}      & \textbf{0.149}      \\
NPBG++ & 0.396               & 0.500               & 0.366               \\
DIBR   & 0.325               & 0.290               & 0.313               \\
    \end{tabular}
    \label{tab:big_eval}    
\end{minipage}
\hfill
\begin{minipage}[t][][t]{0.48\linewidth}
    \centering
        \caption{ Time consumption for different methods. Preprocessing includes training time and the time to generate the proxy geometry. (*FWD only renders at resolution  400$\times$300.) }
    \begin{tabular}{l|ccc}
        Method & Proxy Geometry & Preprocessing & Rendering Time\\ 
        \hline
        Ours & depth maps & < 1\,s & $\sim$ 46\,ms\\ 
        FWD & depth maps & < 50\,s & $\sim$ 84\,ms*\\ 
        SVS & mesh & $\sim$ 4\,h & $\sim$ 3\,s\\ 
        IBRNet & {---} & < 1\,min & $\sim$ 2\,min\\ 
        NPBG++ & point cloud & $\sim$ 1\,h & $\sim$ 100\,ms \\ 
        DIBR & depth maps & < 1\,min & $\sim$  5\,ms\\ 
        ADOP & point cloud & $\sim$ 8\,h & $\sim$ 15\,ms \\ 
        Instant-NGP & volumetric hash & $\sim$ 2\,min & $\sim$ 2\,s \\ 
    \end{tabular}
    \label{tab:time}
\end{minipage}
\hfill
\begin{minipage}[t][][t]{0.23\linewidth}
    \centering
        \caption{ Keyframe selection results on \textsc{Scannet} using frames that passed our selection ({\small s}) or using every 20th view ({\small mod20}) during training. }
    \begin{tabular}{l|c}
        Training Views & LPIPS  $\downarrow$  \\ 
        \hline 
        Ours\textsubscript{s} & 0.2705 \\ 
        Ours\textsubscript{mod20} & 0.3401 \\ 
        SVS\textsubscript{s} & 0.2673 \\ 
        SVS\textsubscript{mod20} & 0.3162 \\ 
    \end{tabular}
    \label{tab:scannet_motion_blur}
\end{minipage}
\end{table*}}

\subsection{Datasets \& Training Procedure}
\label{ssec:data_and_train}

We conducted extensive evaluation utilizing a combination of publicly available datasets, namely the \textsc{Scannet}~\cite{Dai2017}, Redwood~\cite{choiLargeDatasetObject2016} (\textsc{Motorcycle} and \textsc{Sofa}) and \textsc{Tanks and Temples} dataset~\cite{knapitsch2017tanks}, along with our own custom data captured using the Zed 2i camera. 
The Zed 2i employed is a stereo camera equipped with a resolution up to 2K.
Notably, the depth estimation process relies on a state-of-the-art deep learning algorithm~\cite{zed}. 

We have trained the involved networks end-to-end in deferred warping mode (see Sec.~\ref{ssec:deferredwarping}) on a combination of \textsc{Scannet} and \textsc{Tanks and Temples} scenes.
Remarkably, all experiments in the following sections, as well as the supplemental video, were generated using identical network weights.
This shows that our method can generalize well to unseen scenes and even to new modalities like the Zed camera, which was not included in the training dataset.

For more details on training and dataset preparation, we refer to the supplemental material.

\subsection{Live RGB-D Stream Novel View Synthesis}

To evaluate the quality of novel view synthesis during capturing in relation to previous work, we have extended the FWD~\cite{caoFWDRealTimeNovel2022} and Instant-NGP~(INGP)~\cite{mullerInstantNeuralRadiance2022} pipeline to run on growing datasets from our Zed camera.
For a fair comparison, we ran the INGP optimization for the amount of time up to the captured frame.
Our method and FWD do not require a scene specific optimization.
The results are shown in Fig.~\ref{fig:instant_nvs}.
The neural rendering results show that our method is able to produce the sharpest results with the least amount of artifacts.
For example, the colorful wall painting (left) and the world map (right) is clearly recognizable in our approach and heavily blurred for FWD and INGP.

\subsection{Novel View Synthesis on Full Datasets}

\begin{figure*}
\centering
\includegraphics[width=\linewidth]{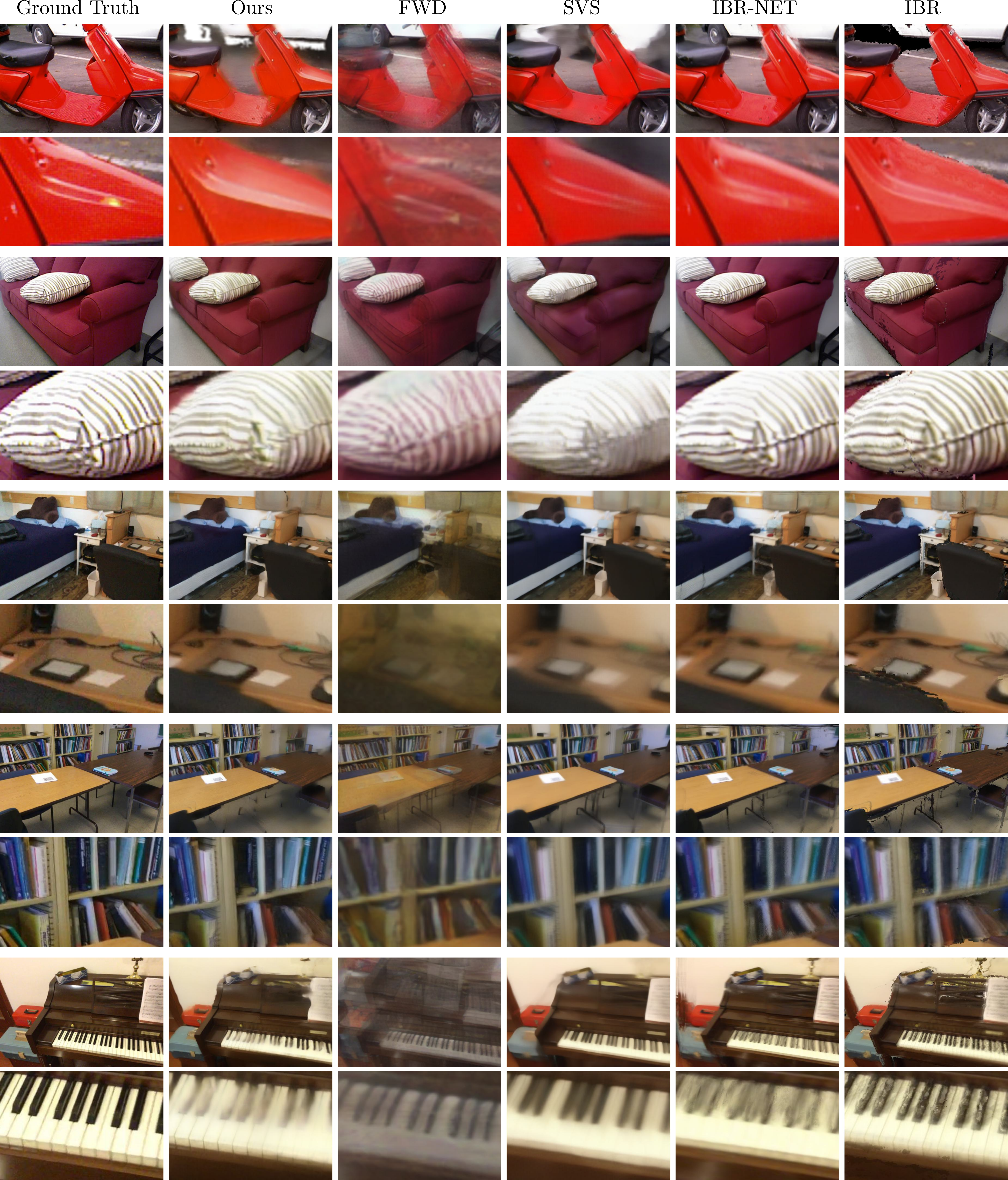} 
\caption{NVS results on \textsc{Motorcycle},  \textsc{Sofa} and \textsc{Scannet}.}
\label{fig:eval_gen}
\end{figure*}

We compare the rendering quality of our approach to state-of-the-art methods that also do not require a scene-specific optimization procedure.
These methods are Stable View Synthesis~(SVS)~\cite{rieglerStableViewSynthesis2021}, IBRNet~\cite{wangIBRNetLearningMultiView2021a}, NPBG++~\cite{rakhimovNPBGAcceleratingNeural2022}, FWD~\cite{caoFWDRealTimeNovel2022}, and DIBR (a depth image-based rendering method similar to FragmentFusion~\cite{ruckertFragmentFusionLightWeightSLAM2019}).
Fig.~\ref{fig:eval_gen} shows some exemplary results.
In Tab.~\ref{tab:big_eval}, we summarize the measured image quality of the produced unseen views evaluated using the LPIPS~\cite{zhangUnreasonableEffectivenessDeep2018} perceptual loss.
Our approach reaches a similar LPIPS score compared to the state-of-the-art, while being significantly more efficient without any preprocessing required.

\subsection{Preprocessing and Render Time}

In Tab.~\ref{tab:time}, we provide an overview of preprocessing and rendering time for ours and related work for medium-sized scenes, e.g. rooms or buildings.
SVS and NPBG++ require an initial point cloud or mesh as a geometric proxy. 
Using COLMAP, this takes 2 - 6 hours.
Additionally, several preprocessing steps have to be conducted before rendering can take place.
For example, INGP needs a training stage to optimize the hash grid.
Our approach, requires only minimal preprocessing and can be operated on live RGB-D streams.
Rendering a $1296\times968$ image takes around 46\,ms with our method.
The other systems require seconds to minutes.
A detailed benchmark for differently sized scenes is presented in Tab.~\ref{tab:time2}.
Note that the high inference time of the encoder is hidden in the average total render time as our caching reduces the number of encodings drastically. Even though render times seem competitive, note that warping all available feature maps per frame relies on an excessively sized cache that can store the feature maps of all available source view.

\begin{table}[]
\footnotesize
    \caption{ Benchmarks of our pipeline (1296$\times$968\,px , Nvidia RTX 2080Ti, averaged over 256 frames for a moving camera). The average total frame time includes view selection (if used), encoding of uncached views, (a separate warping pass of all depth maps for the deferred mode), warping of (selected) feature maps, and decoding.}
\begin{tabular}{l|cc}
Total Number of Source Views Available                                                  & 50 Views & 207 Views \\
\hline\hline
Inference Encoder                                                                               & 35\,ms       & 35\,ms        \\

Inference Decoder                                                                               & 12\,ms       & 12\,ms        \\
\hline\hline
\begin{tabular}[c]{@{}l@{}}Avg Total Time: Forward\\ 15 selected Views for Features \& 15 for Depth \end{tabular}  & 46\,ms       & 67\,ms        \\
\hline
\begin{tabular}[c]{@{}l@{}}Avg Total Time: Deferred\\ 15 selected Views for Features \& all for Depth\end{tabular}  & 82\,ms       & 160\,ms       \\
\hline
Including View Selection                                                                          & 7\,ms        & 26\,ms        \\
\hline\hline
\begin{tabular}[c]{@{}l@{}}Avg Total Time: Forward\\ all Views for Features \& all for Depth\end{tabular}          & 66\,ms       & 131\,ms       \\
\hline
\begin{tabular}[c]{@{}l@{}}Avg Total Time: Deferred\\ all Views for Features\& all for Depth \end{tabular}         & 83\,ms       & 168\,ms      \\
\hline
Without View Selection                                                                          & -        & -        \\
\end{tabular}
    \label{tab:time2}
\end{table}

\subsection{Ablation Studies}

\subsubsection{Keyframe Selection}

\begin{figure}
\centering
\includegraphics[width=\linewidth]{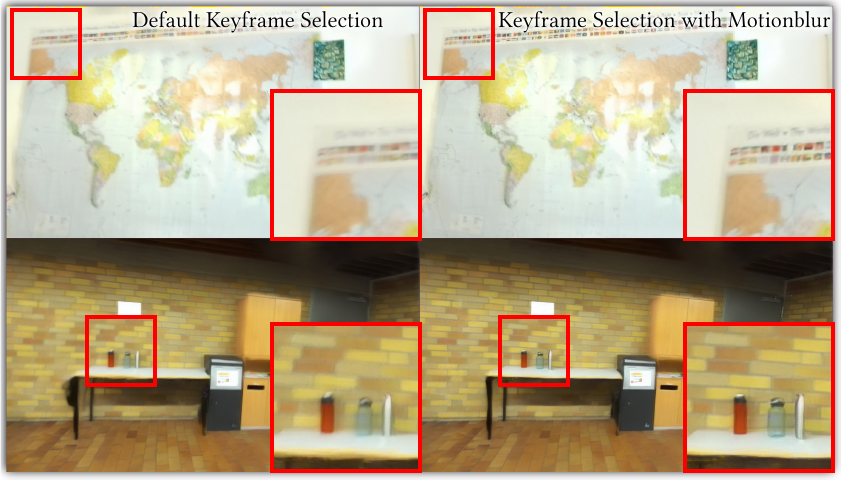}
\caption{ 
Left: neural rendering with default keyframe selection. Right: our improved keyframe selection.}
\label{fig:motion_blur}
\end{figure}

We incorporate an effective motion-blur detection and compensation scheme by removing blurry frames from the input stream (see Sec.~\ref{ssec:keyframes}). 
In Tab.~\ref{tab:scannet_motion_blur}, we evaluate our improved keyframe selection on the \textsc{Scannet} dataset for our approach and SVS~\cite{rieglerStableViewSynthesis2021}.
Both methods profit from our keyframe selection and show a significantly improved LPIPS score,
thus, indicating that also methods with extensive preprocessing and rendering~(see Tab.~\ref{tab:time}) can benefit from our selection. 
A visual comparison on two indoor scenes is presented in Figure \ref{fig:motion_blur}.

\subsubsection{View Extrapolation}
\begin{figure}
    \centering
    \includegraphics[width=\linewidth]{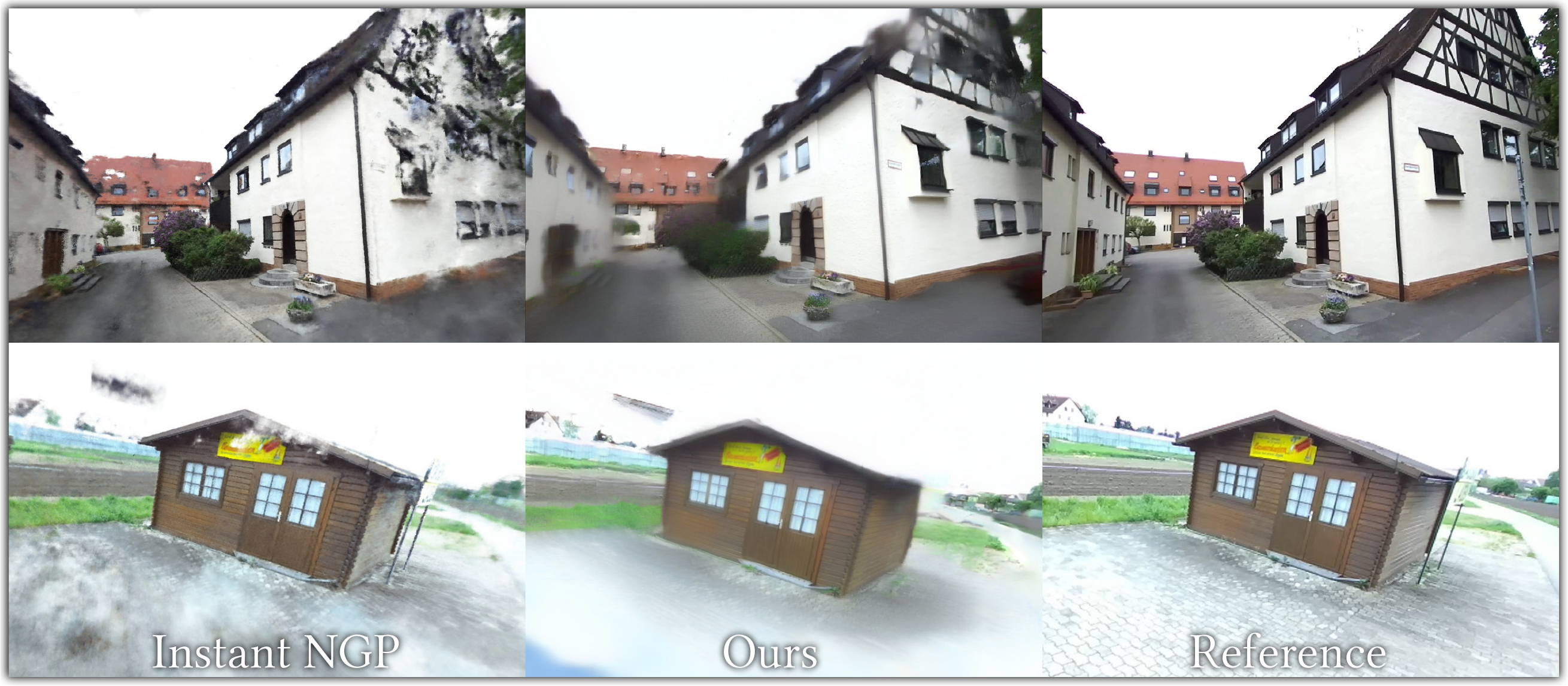}
    \caption{ View extrapolation results for INGP (left) and our method (center). Note that there are severe cloud-like artefacts in the INGP rendering, while ours gets close to the reference (right).}
\label{fig:view_extrapolation}
\end{figure}

Figure \ref{fig:view_extrapolation} shows a synthesized novel view for a virtual camera that is far away from the input trajectory.
Volumetric-based methods like INGP introduce cloud-like artifacts in this case. 
Our approach is able to render a clean image with only few noticeable inaccuracies.

\subsubsection{Loop Closure}
\begin{figure}
    \centering
    \includegraphics[width=\linewidth]{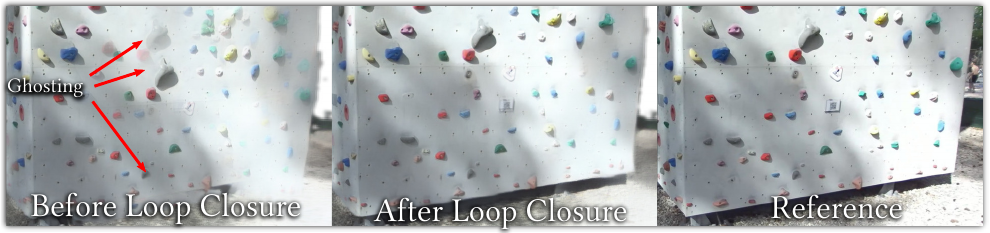}
    \caption{
        Tracking a loop can cause ghosting artifacts in the rendering due to accumulated drift (see left image).
        The SLAM system automatically detects and closes these loops.
        Our rendering immediately reflects these changes, since no global model is optimized (center image).
    }
\label{fig:loop_closure}
\end{figure}

An advantage of LiveNVS is that the poses of previous keyframes can be updated at any time.
Thus, our renderings can immediately mirror detected loops of the tracking system.
Fig.~\ref{fig:loop_closure} shows our rendering right before and after a loop closure.
The ghosting artifacts due to miss-aligned poses before the loop closure are removed on the right.

\subsubsection{Source of depth maps}

\begin{figure}
\centering    

    \begin{minipage}[t]{0.325\linewidth}
    \includegraphics[width=\linewidth, height=2cm]{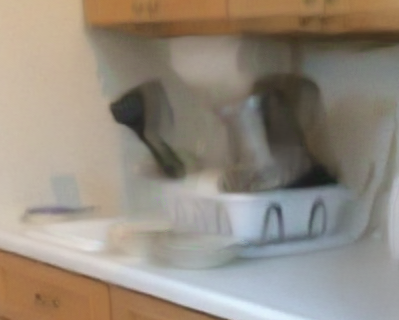}%
    \end{minipage}
    \hfill
    \begin{minipage}[t]{0.325\linewidth}
    \includegraphics[width=\linewidth, height=2cm]{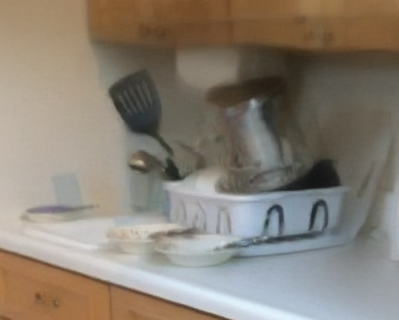}%
    \end{minipage}    
    \hfill
    \begin{minipage}[t]{0.325\linewidth}
    \includegraphics[width=\linewidth, height=2cm]{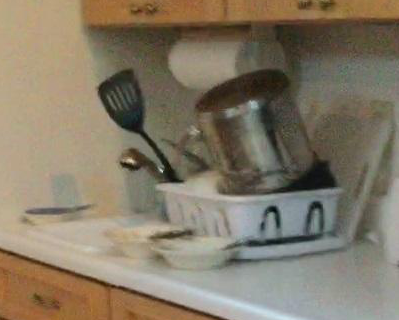}%
    \end{minipage}
    
    \caption{ Crops of renderings using depth maps from active sensor~(left) vs.
    mesh~(center). Groundtruth(right).}
    \label{fig:depth_vs_mesh}
   
\end{figure}

Alternatively to depth maps from active sensors, we can also use depth maps rendered from offline reconstructed meshes.
This allows us to profit from subsequent post-processing of the original depth maps or other offline methods delivering high quality 3D reconstructions.
Especially, in areas where depth maps from active sensors are prone to be noisy or erroneous, visual fidelity can be improved, as shown in Fig.~\ref{fig:depth_vs_mesh}.
However, on the \textsc{Scannet} dataset a quantitative evaluation shows only a minor improvement of the LPIPS score from $0.271$ to $0.269$.

\subsubsection{View Selection}
\begin{figure}     
    \centering
    \includegraphics[width=\linewidth]{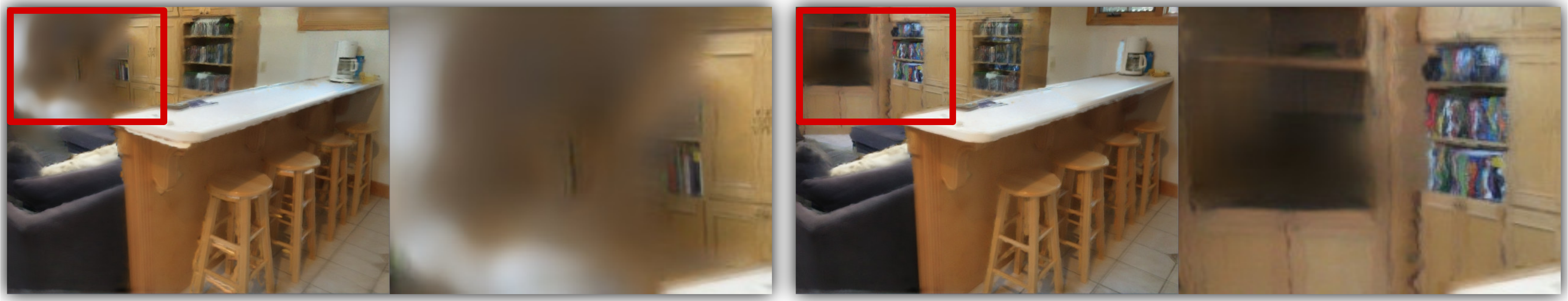}%

    \caption{ Renderings using untiled~(left) and tiled~(right) view selection.}
    \label{fig:view_selection}
\end{figure}

As described in Sec.~\ref{ssec:view_selection}, we do a coverage-based view selection where we tile the target image frame.
Fig.~\ref{fig:view_selection} visualizes the benefit of such a selection scheme as big holes in the background are avoided effectively in this view.

\section{Limitations and Future Work}
\label{sec:limitations}

During our evaluation, we found a few limitations of our work.
The major limitation becomes visible if the pose estimates are extremely inaccurate or the depth maps are too noisy.
In this case, our rendering network gets inconsistent feature descriptors and outputs a blurred image of the scene.
Our surface fusion algorithm might fail if the noise of the depth values exceeds the threshold~(see $\Delta(d)$ in Eq.~\ref{eq:w_d}).
Other approaches, which rely on inverse rendering optimization, can correct such input errors and output sharper renderings.
However, this optimization step is often time consuming, while ours can be directly applied to live video stream.

The second limitation is a potential temporal instability when the view selection swaps out an input image for another, resulting in flickering.
This flickering is especially visible if exposure time or white balance vary throughout the stream.
The temporal feedback loop incorporated in our decoder alleviates this issue to a certain extent but a too high contribution of previous frames to counter extreme flickering may lead to ghosting. 
In such cases, we see the warping of the previous feature maps as a promising extension.

Since both limitations are a result of imperfect input data, a combination of our approach with an inverse rendering back-end might solve these issues.
The idea would be to run the inverse optimization in a secondary thread on the previous keyframes to refine the pose, depth map, exposure time and white balance.
Our pipeline can then use the optimized keyframes to synthesize high-quality real-time rendering of the scene.

\section{Conclusion}
\label{sec:conclusion}

In this work, we have proposed a novel pipeline for real-time novel view synthesis on RGB-D video streams.
The core of our pipeline is the flexible neural fusion function, which amalgamates the color and depth data of nearby keyframes in image space of the target view on-the-fly.
The major advantage of our approach is that it requires zero preprocessing or training time on new scenes and  can be readily applied on consumer hardware.
During an extensive evaluation, we have shown that the rendering quality matches or outperforms other generalizing neural rendering approaches even if they relied on a time-consuming preprocessing on the dataset.

\begin{acks}

We would like to thank all members of the Visual Computing Lab Erlangen for their support and fruitful discussions.
Specifically, we appreciate Mathias Harrer's contribution to the evaluation and Dominik Penk's help regarding the dataset preparation. 
We also thank Ashutosh Mishra for his insights about prior arts.

The authors gratefully acknowledge the scientific support and HPC resources provided by the National High Performance Computing Center  of the Friedrich-Alexander-Universit\"at Erlangen-N\"urnberg (NHR@FAU) under the project \textit{b162dc}. NHR funding is provided by federal and Bavarian state authorities. NHR@FAU hardware is partially funded by the German Research Foundation (DFG) – 440719683.
Linus Franke was supported by the Bavarian Research Foundation (Bay. Forschungsstiftung) AZ-1422-20.
Joachim Keinert was supported by the Free State of Bavaria in the DSAI project.

\end{acks}

\bibliographystyle{ACM-Reference-Format}
\bibliography{bibliography}

\clearpage

\appendix
{\centering
\Huge \textbf{Supplemental Material}}
\vspace{.2cm}

In this supplementary material, we provide extra ablation studies and details on experiments and implementation.

\section{Implementation details} 

\subsection{Camera Pose Estimation}
For the purpose of tracking, we employed Snake-SLAM~\cite{9476760}, for both the Redwood and Zed datasets. 
The evaluation on Scannet was done using the provided camera parameters and for Tanks and Temples we used COLMAP~\cite{schonbergerPixelwiseViewSelection2016} reconstructions.

\subsection{Details on Training}
The used dataset is a combination of 82 Scannet scenes and 14 Tanks and Temples scenes.
In each iteration, eight views are drawn from the list provided by the view selection algorithm and sampled according to pre-computed per-pixel lists.
These views are then warped and fused into a target view.
As loss we solely used the VGG loss~\cite{zhangUnreasonableEffectivenessDeep2018}, computed from predicted target frame and reference.
In total, we run this training for 500K steps with a batch size of 1 and learning rate $10^{-4}$ using the Adam optimizer. This took ~24h on an RTX 3090 GPU.

\subsection{Temporal Feedback of the Decoder}
During inference, intermediate feature maps of the current and the previous frame are blended according to a blending factor within [0, 1] that can be set interactively.
To allow for this feature, we randomly sampled its value during training. 
In all of our evaluation experiments, we set this blending factor to 0.1.

\section{Further Evaluation}

\subsection{Weighting Scheme}

\begin{figure}[b]
\centering
\includegraphics[width=.9\linewidth]{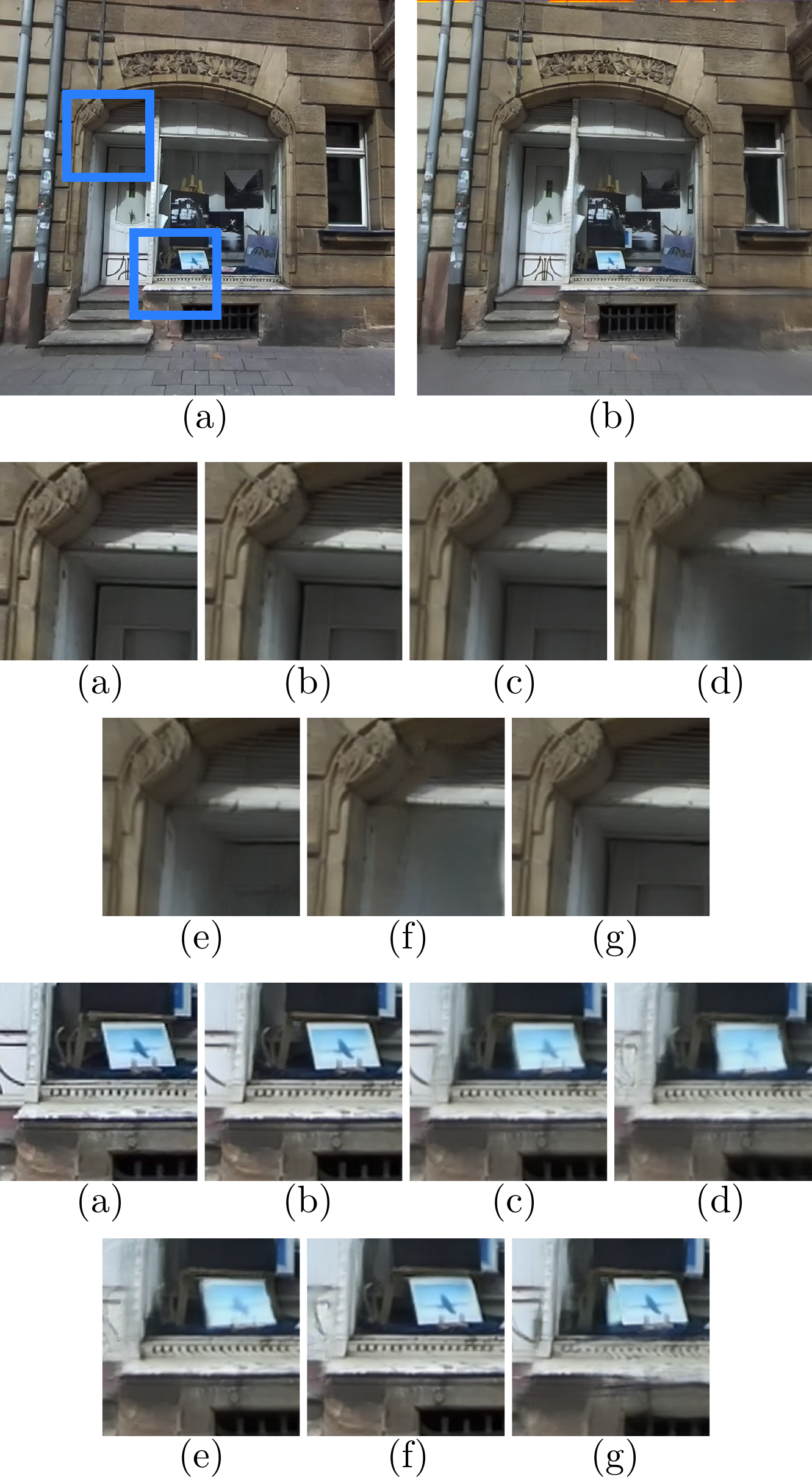}

\caption{Influence of weighting scheme. (a)~Reference, (b)~$w=(w_d w_v w_i)^5$, (c)~$w=w_d w_v w_i$, (d)~$w=1$, (e)~$w=w_v$, (f)~$w=w_d$, (g)~$w=w_i$.}
\label{fig:weights}
\end{figure}

Figure~\ref{fig:weights} illustrates the influence of the involved weights on the output image.

\subsection{Auxiliary Features}

We evaluated different pipeline variations which mainly focused on providing additional in- and outputs of the en- and decoding networks to improve quality.
We did not incorporate additional networks like MLPs for feature translation~\cite{rieglerStableViewSynthesis2021} or transformers~\cite{wangIBRNetLearningMultiView2021a} as done in related work to keep the pipeline as light-weight as possible.

The simplest one is using RGB as input for the encoder only. 
Additionally, we may add depth~(+D) as additional feature, where +D$_o$ means the original depth from the depth camera was provided as input for the encoder and the fused depth of the target view was provided for the decoder. 
A fused depth map of the source images was provided for the encoder for the variant +D$_f$ (which is only possible when backwards warping).
+V indicates that we additionally compute per-pixel feature variance of the target feature map.
The variant +$c$ lets the encoder output an additional channel which is used as weight during the fusion phase and expresses confidence of predicted neural features. 

The Table~\ref{tab:variants} shows mean results for the evaluation scenes of the \textsc{Scannet} dataset.
Most variations performed very similar and are within a 1\,\% margin.
SSIM was not sensitive to image differences at all, while L1 showed up to 2.5\,\% relative difference between the best and worst performing variant.

In general, we can observe slightly improved metric values, less blurring and artifact reduction when adding input features to the networks.
We identify RGB+D$_f$+$c$ as best compromise between image quality improvements and investment of additional compute time and memory consumption.

\begin{table}[]
    \centering
    \begin{tabular}{l|l|l|l|l}
    Variant & PSNR $\uparrow$ & L1 $\downarrow$ & SSIM  $\uparrow$ & LPIPS  $\downarrow$\\
    \hline\hline
        RGB & 22.44 & 0.1428 & \textbf{0.7967} & 0.4016 \\ 
        RGB+$c$ & 22.48 & 0.1422 & \textbf{0.7976} & \textbf{0.3986} \\ 
        RGB+V & \textbf{22.51} & 0.1428 & \textbf{0.7974} & 0.4032 \\ 
        RGB+D$_f$ & \textbf{22.60} & 0.1399 & \textbf{0.7981} & 0.4019 \\ 
        RGB+D$_f$+$c$ & \textbf{22.57} & \textbf{0.1392} & \textbf{0.7972} & 0.4007 \\ 
        RGB+D$_f$+V & \textbf{22.55} & 0.1405 & \textbf{0.7973} & 0.4010 \\ 
        RGB+D$_o$+V & 22.47 & 0.1411 & \textbf{0.7953} & \textbf{0.3987} \\ 
        RGB+D$_f$+$c$+V & 22.46 & 0.1427 & \textbf{0.7962} & 0.4011 \\
    \end{tabular}
    \caption{Performance of various network variants. Bold entries performed best within 0.5\% relative difference. }
    \label{tab:variants}
\end{table}

\subsection{Additional Results}

\begin{table*}[ht]
    \centering
    \caption{Quantitative comparison of the rendering quality. Note, that our approach is the only method that directly processes the live RGB-D stream, while all other approach use the globally fused point cloud or mesh.}
    \begin{tabular}{l|cccc|cccc|cccc}
        ~ & \textsc{Scannet} & ~ & ~ & ~ & \textsc{Sofa} & ~ & ~ & ~ & \textsc{Motorcycle} & ~ & ~  \\ 
        ~ & PSNR $\uparrow$ & L1 $\downarrow$ & SSIM  $\uparrow$ & LPIPS  $\downarrow$ & PSNR $\uparrow$ & L1 $\downarrow$ & SSIM  $\uparrow$ & LPIPS  $\downarrow$ & PSNR $\uparrow$ & L1 $\downarrow$ & SSIM  $\uparrow$ & LPIPS  $\downarrow$ \\ 
        \hline
        \hline
        Ours & 20.40 & 0.061 & 0.747 & 0.274 &      17.15 & 0.093 & 0.594 & 0.303 &     19.48 & 0.082 & 0.714 & 0.234 \\ 
        FWD & 17.27 &  0.106 & 0.695 & 0.428 &      15.27 & 0.121 & 0.488 & 0.451 &     19.87 & 0.072 & 0.690 & 0.295 \\
        SVS & 23.76 & 0.042 & 0.796 & 0.316 &       20.01 & 0.061 & 0.694 & 0.247 &     22.64 & 0.051 & 0.744 & 0.237 \\ 
        IBRNet & 24.96 & 0.035 & 0.802 & 0.248 &    23.56 & 0.041 & 0.758 & 0.179 &     27.77 & 0.028 & 0.821 & 0.149 \\ 
        NPBG++ & 20.90 & 0.066 & 0.758 & 0.396 &    15.34 & 0.125 & 0.535 & 0.500 &     17.40 & 0.103 & 0.654 & 0.366 \\ 
        DIBR & 19.62 & 0.065 & 0.708 & 0.325 &      16.69 & 0.091 & 0.592 & 0.290 &     18.28 & 0.084 & 0.662 & 0.313 \\ 
    \end{tabular}
    \label{tab:big_eval}
\end{table*}

\begin{table*}[ht]
    \centering
    \caption{Quantitative comparison of the rendering quality for finetuned~(\textsubscript{ft}) experiments.\label{tab:big_eval_ft}}
    \begin{tabular}{l|cccc|cccc|cccc}
        ~ & \textsc{Scannet} & ~ & ~ & ~ & \textsc{Sofa} & ~ & ~ & ~ & \textsc{Motorcycle} & ~ & ~  \\ 
        ~ & PSNR $\uparrow$ & L1 $\downarrow$ & SSIM  $\uparrow$ & LPIPS  $\downarrow$ & PSNR $\uparrow$ & L1 $\downarrow$ & SSIM  $\uparrow$ & LPIPS  $\downarrow$ & PSNR $\uparrow$ & L1 $\downarrow$ & SSIM  $\uparrow$ & LPIPS  $\downarrow$ \\ 
        \hline
        \hline
        Ours\textsubscript{ft} & 21.15 & 0.055 & 0.758 & 0.260 & 19.59 & 0.077 & 0.716 & 0.200 & 17.36 & 0.092 & 0.618 & 0.247 \\ 
        SVS\textsubscript{ft} & 23.26 & 0.044 & 0.785 & 0.260 & 25.04 & 0.038 & 0.775 & 0.162 & 20.11 & 0.060 & 0.688 & 0.206 \\ 
        IBRNet\textsubscript{ft} & 25.42 & 0.033 & 0.804 & 0.229 & 28.14 & 0.027 & 0.812 & 0.163 & 23.59 & 0.042 & 0.743 & 0.182 \\ 
        NPBG++\textsubscript{ft} & 21.66 & 0.058 & 0.763 & 0.363 & 17.02 & 0.111 & 0.637 & 0.392 & 15.26 & 0.122 & 0.521 & 0.496 \\ 
        ADOP & 24.78 & 0.039 & 0.694 & 0.208 & 24.26 & 0.044 & 0.598 & 0.172 & 20.34 & 0.066 & 0.585 & 0.236 \\ 
    \end{tabular}
\end{table*}

\begin{figure*}
\centering
\includegraphics[width=\linewidth]{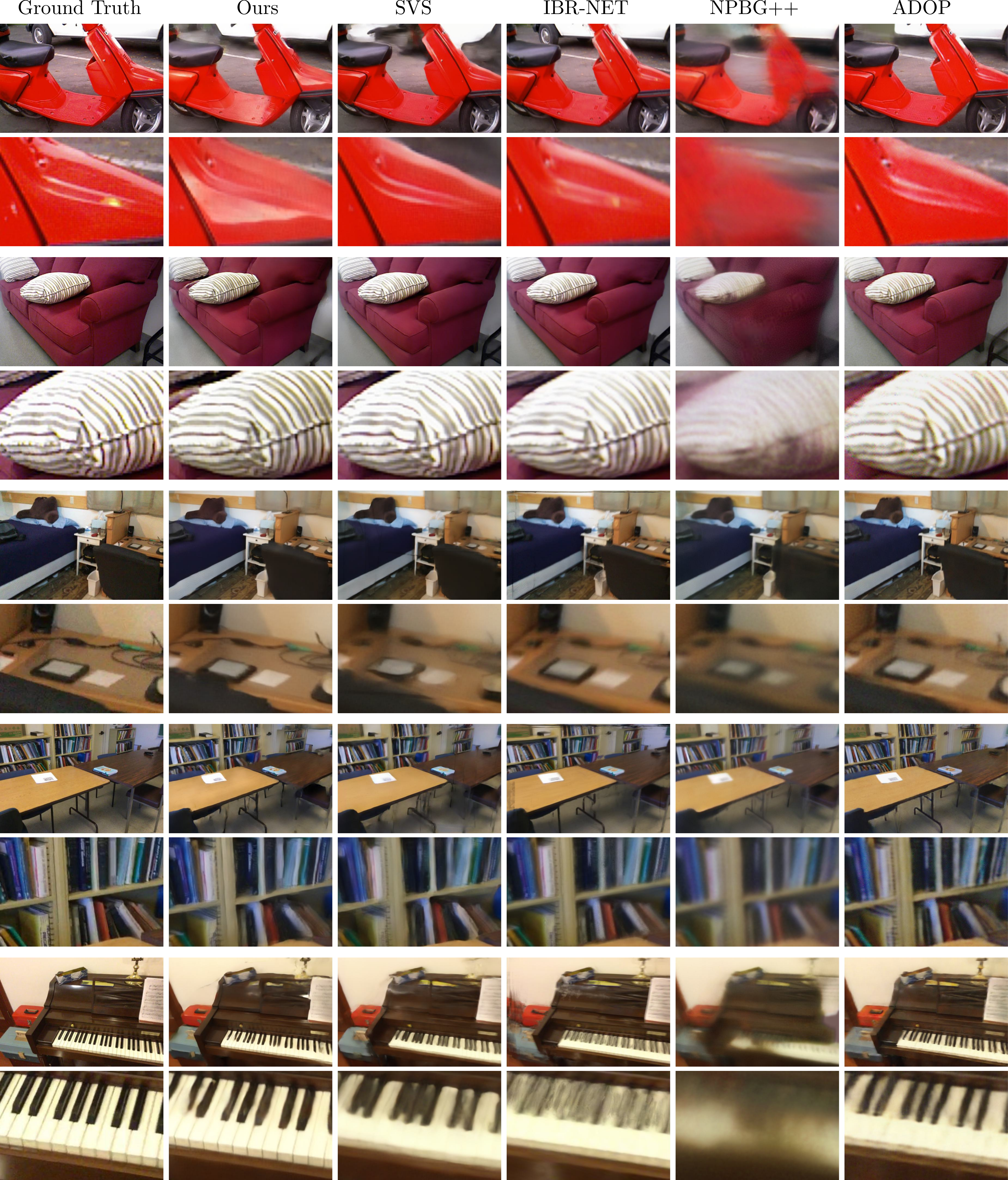} 
\caption{\textbf{Finetuned} NVS Results on \textsc{Motorcycle},  \textsc{Sofa} and \textsc{Scannet}.}
\label{fig:eval_ft}
\end{figure*}

Table~\ref{tab:big_eval} show additional metrics to the experiment conducted in the paper. 
Additionally, we performed some tests regarding scene-specific fine-tuning. 
Compared to methods which are tailored to scene-specific reconstruction (often needing hours per scene or multiple minutes per view), metrics indicate only mid-range results, see Table~\ref{tab:big_eval_ft}.
However, Figure~\ref{fig:eval_ft} illustrates that synthesized views generally look plausible.

\end{document}